    \documentclass[letterpaper, 10 pt, conference]{ieeeconf}  

    \IEEEoverridecommandlockouts                              
    
    \usepackage{graphicx}
    \usepackage[table,xcdraw]{xcolor}
    \usepackage{subcaption}
    \usepackage{amsmath}
    \usepackage{multirow}
    \usepackage{hyperref}
    
    \newlength{\subfigwidth}
    \newlength{\subfigwidthlarge}
    \title{
    \small{\textcolor{gray}{This paper has been accepted for publication at the the IEEE/RSJ International Conference on Intelligent Robots and Systems (IROS), Detroit, 2023. ©IEEE}}
    \vspace{10pt}
    \\
    \LARGE \bf
    Depth self-supervision for single image novel view synthesis}

    \author{Giovanni Minelli$^{1}$ \hspace{0.7cm} Matteo Poggi$^{1}$ \hspace{0.7cm} Samuele Salti$^{1}$
    \thanks{$^{1}$University of Bologna}
    }

\begin{document}

    \maketitle
    \thispagestyle{empty}
    \pagestyle{empty}

    \begin{abstract}   
    In this paper, we tackle the problem of generating a novel image from an arbitrary viewpoint given a single frame as input. While existing methods operating in this setup aim at predicting the target view depth map to guide the synthesis, without explicit supervision over such a task, we jointly optimize our framework for both novel view synthesis and depth estimation to unleash the synergy between the two at its best. Specifically, a shared depth decoder is trained in a self-supervised manner to predict depth maps that are consistent across the source and target views.
    Our results demonstrate the effectiveness of our approach in addressing the challenges of both tasks allowing for higher-quality generated images, as well as more accurate depth for the target viewpoint.
    \end{abstract}

\section{Introduction}
In many fields, data is a necessary burden. It plays a crucial role in the industry, helping in decision-making, progress monitoring, and gaining insights. It becomes essential in applications involving deep learning, such as most computer vision tasks.
However, obtaining data in large quantities is often challenging, specifically when physical space is limited and does not allow for deploying multiple sensors, as often happens in robotic systems, or when cost and time are significant constraints. Ongoing research is trying to discover alternative ways to enable realistic data generation to compensate for this lack. Among them, we can find Novel View Synthesis (NVS) in computer vision, as the task of generating images framing unseen or occluded parts in a scene. It can be used for applications such as image editing, enhancement of visual experiences or virtual reality \cite{MartinBrualla2018LookinGoodEP}, where it would be desirable to generate new frames on the fly out of a scene, or even in robotic navigation, potentially allowing an agent to infer and predict a dangerous situation before it could happen \cite{XinYangFastDepthObsAvoid}.
    
To tackle the NVS task, it is generally necessary to reason about the 3D structure of the scene, which enables to infer the relative motion of objects represented under a view transformation. While recent methods have made significant progress using multiple views to reconstruct 3D scene geometry \cite{Mildenhall2020NeRFRS, mueller2022instant, Kopf2013ImagebasedRI, Rematas2017NovelVO, Seitz2006ACA, Fitzgibbon2003ImageBasedRU, Ladicky2014PullingTO}, they cannot be seamlessly extended to work with a single image.
When working with a single 2D image, estimating the underlying geometry of the scene becomes an ill-posed problem. Additional prior information in the form of depth \cite{Wiles2020SynSinEV, Liu2018GeometryAwareDN, Zhu2018ViewEO} or light fields \cite{Srinivasan2017LearningTS} can be used to aid in this estimation and ensure consistency in the geometric structure. However, acquiring such additional data is challenging and requires ad-hoc sensors.

 \begin{figure}
     \centering
     \includegraphics[clip, width=0.48\textwidth]{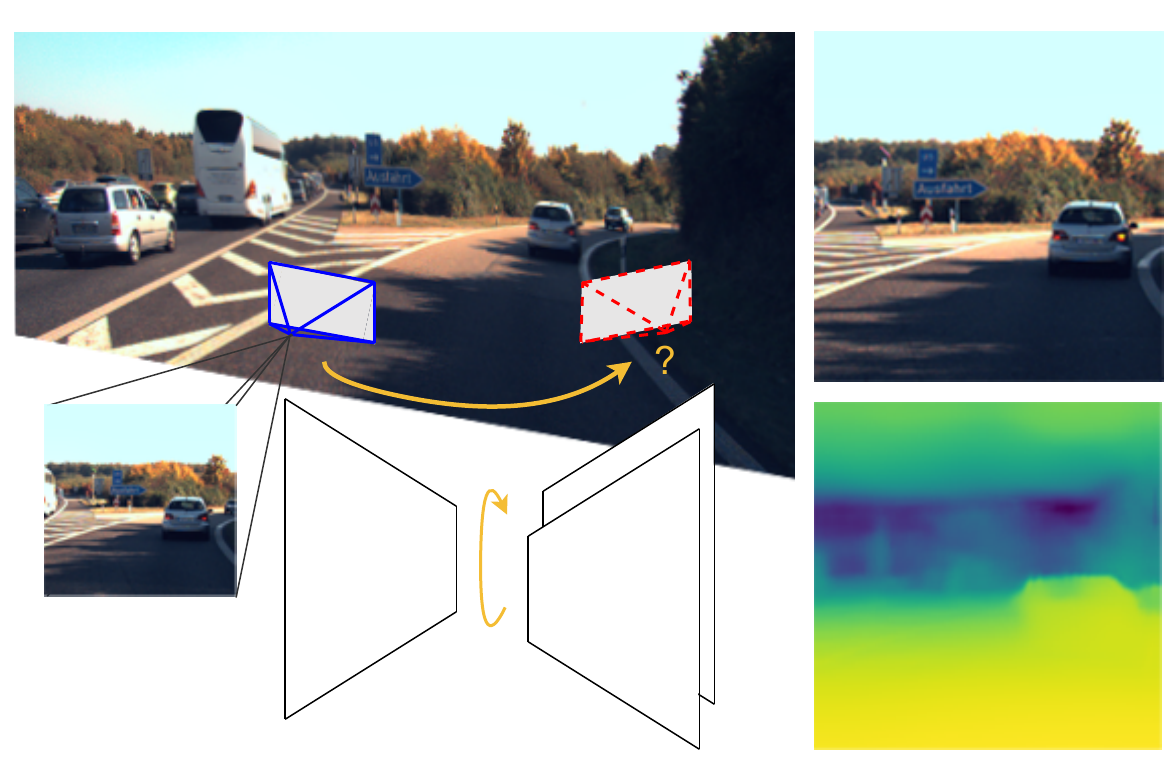}
     \caption{\textbf{Framework overview.} Our model generates a novel, target image and its depth map out of a single, source frame.}
     \label{fig:teaser}
 \end{figure}
 
To relax these constraints, recently some methods \cite{chen2019mono,Hou_2021_WACV} proposed to implicitly encode knowledge about the scene geometry in a compact latent embedding extracted from the source image and to reconstruct the desired novel view by transforming such a representation according to the relative pose between the two viewpoints. We dub them \textit{source-to-target} approaches. Although such a paradigm is appealing, some limitations dampen its overall effectiveness, where existing frameworks learn the NVS task with direct supervision being limited to the image generation process itself. However, the synthesis process is a direct consequence of the scene geometry which is implicitly modeled during inference -- e.g., in the form of depth maps \cite{chen2019mono,Hou_2021_WACV} -- yet never explicitly optimized. We argue this lack of supervision on geometry yields sub-optimal results, dampening the great potential of source-to-target approaches.
In this paper, we propose a new pipeline for source-to-target NVS that explicitly reasons on a latent representation of the scene at the geometry level.
The key point of our approach consists in exploiting self-supervised depth estimation to better guide the network in learning such underlying geometry, and to allow for the proper generation of novel views. The synthesis task is performed by two decoders, parsing the encoded embeddings, correcting the distortion effects and producing highly detailed output, as those sketched in Figure \ref{fig:teaser}. The synergy between NVS and depth estimation yields higher quality results in comparison to existing approaches \cite{chen2019mono,Hou_2021_WACV}, both on synthetic images and on real data, for both generated novel images and estimated depth maps. Our code is available at \url{https://github.com/johnMinelli/TwoWaySynth/}.
    
\section{Related Work}
    
In this section, we review the research trends in NVS being most relevant to our work.

\subsection{View synthesis from multi-view images.}
This family of approaches can be used in a variety of applications, including 3D reconstruction, object recognition, and scene understanding. It can be performed by collecting images from multiple cameras or by taking multiple pictures from different angles with a single camera and then reconstructing the 3D structure of the scene by exploiting the consistency between views.
Traditionally, this was done using depth maps \cite{Chaurasia2013DepthSA}, or multi-view geometry methods \cite{Debevec1996ModelingAR, Kopf2013ImagebasedRI, Rematas2017NovelVO, Seitz2006ACA, Shade1998LayeredDI, Fitzgibbon2003ImageBasedRU, Ladicky2014PullingTO}, but those approaches often suffer from unreliable photometric consistency and artifacts in the reconstructed images. 
Then, the use of neural networks -- CNNs in particular -- has become a popular approach for NVS, by exploiting deep features instead of using explicit images. This strategy leverages geometrical and optical properties such as depth and occlusions to generate novel views \cite{Penner2017Soft3R, Hedman2018DeepBF, Zhou2018StereoML, Novotn2019PerspectiveNetAS, Aliev2020NeuralPG, MartinBrualla2018LookinGoodEP}. Furthermore, the fully differentiable frameworks often involved allows for merging geometry estimation and novel view synthesis for better results. \cite{Sun2018MultiviewTN} proposed a framework to combine flow-based predictions from multiple input views and then do pixel generation prediction via confidence maps. \cite{Shi2021SelfSupervisedVL} reasons about the visibility of the pixels in the different images to implement a consensus volume and then determine the depth of the scene usable to warp source pixels into target views. \cite{Cao2022FWDRN} employs Transformers to fuse 3D point clouds relative to the target viewpoint, extracted from a sparse set of input images with estimated depths.

A very popular and novel paradigm for NVS from multiple images is Neural Radiance Fields (NeRF) \cite{Mildenhall2020NeRFRS}, using MLPs to infer the volume density and view-dependent emitted radiance from 5D input coordinates -- spatial locations and viewpoint directions.
Images are generated by querying the MLPs for a set of 3D points along the camera ray of each pixel, through volumetric rendering.
As a downside, NeRF necessitates millions of queries to the MLP network and does not generalise across different scenes -- i.e., it requires per-scene training. 
Some variants \cite{Yu2021pixelNeRFNR, Chen2021MVSNeRFFG, Wang2021IBRNetLM, Trevithick2021GRFLA} address the generalisation issue of NeRF showing comparable performance on selected testing scenes, while others focus on training and rendering speed \cite{sun2021direct,yu2022plenoxels,mueller2022instant}

The underlying 3D geometry of the scene is crucial to properly render novel views. A variety of representations (both implicit and explicit) have been used for NVS. Some methods approaching NVS from multi-view images estimate it by means of 3D representations like voxels \cite{Tulsiani2018FactoringSP, Choy20163DR2N2AU, Sitzmann2019DeepVoxelsLP, Guo2022FastAE} or meshes \cite{Riegler2020FreeVS, Riegler2021StableVS, Jena2022NeuralMG, Hu2021WorldsheetWT}, but these methods can be computationally expensive. Some approaches \cite{Riegler2020FreeVS, Riegler2021StableVS} obtain photorealistic results using a mixed approach: starting from a possibly incomplete depth estimation obtained through a structure-from-motion method \cite{Schnberger2016StructurefromMotionR}, they build a point cloud used as starting point for the meshing process. Other works with higher computational and memory efficiency rely instead on point clouds directly \cite{Wiles2020SynSinEV, Aliev2020NeuralPG, Cao2022FWDRN, Rockwell2021PixelSynthGA}. However, point-based representations can yield artifacts or holes between points after projection on the image plane, unless complex generative methods are included in the pipeline to fill the gap with plausible generated content \cite{Wiles2020SynSinEV, Meshry2019NeuralRI, Tsutsui2022NovelVS}.

All the approaches introduced so far obtain impressive NVS results, yet require multiple images as input, a hard-to-meet constraint in most applications. In contrast, our method only requires a source image to render a novel view.
    
\subsection{Source-to-target view synthesis}

A different approach to NVS aims at generating a target novel view given a single source image.
The lack of geometry knowledge due to the absence of multiple images is compensated by exploiting additional cues, often in the form of ground truth geometry or semantic information to train the 3D representation \cite{Niklaus20193DKB, Shin20193DSR, Tulsiani2018FactoringSP}. We mention in particular the use of depth maps \cite{Wiles2020SynSinEV, Liu2018GeometryAwareDN, Zhu2018ViewEO}, surface normals \cite{Liu2018GeometryAwareDN} and light field images \cite{Srinivasan2017LearningTS}. However, collecting large quantities of such data in real-world settings can be challenging. As a result, synthetic environments are often used for training, not excluding the possibility of later refining the results using real-world data. 

Image-based rendering techniques use geometry information to generate new views using the pixels from the source images. By projecting one or more images onto a target view and blending the results, these methods can generalize to unseen data creating free-viewpoint images \cite{Choi2019ExtremeVS, Penner2017Soft3R, Hedman2018DeepBF}.
\cite{Park2017TransformationGroundedIG} performs a transformation on the 3D latent space of an encoder–decoder network used to predict an occlusion-aware flow and then refine the transformed image with a completion network. \cite{Bauer_2021} shows how depth estimation and novel view synthesis tasks are tightly linked by generating new data to supervise the first task. \cite{chen2019mono} obtain novel views using inverse warping in target position by using a 3D transforming autoencoder \cite{Hinton2011TransformingA} to predict the depth map needed. Generally, directly generated images may suffer from blurriness, lack of texture details or inconsistency of identity, and \cite{Hou_2021_WACV} solve that by introducing a decoding step to obtain the NVS output. 

Among these methods, \cite{chen2019mono,Hou_2021_WACV} allows for generating target views without any additional supervision, by implicitly modeling the geometry of the scene in a compact latent representation, that is transformed according to the relative camera poses between the source and the target images to generate the latter. However, these frameworks are supervised at the image level only, without any direct optimization of the latent representation.     
Our work shares the main objective with these approaches, yet overcomes this latter limitation by explicitly supervising the process at the geometry level by means of self-supervised depth estimation.

    \begin{figure*}
        \centering
        \includegraphics[height=4.6cm]{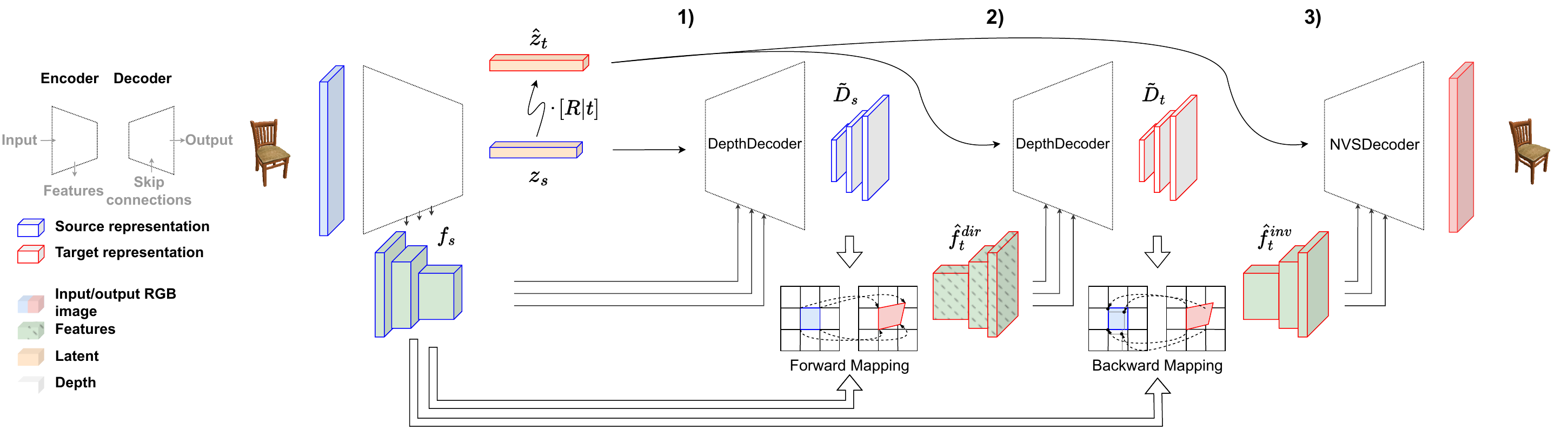}
        
        \caption{\textbf{Proposed architecture for joint NVS and depth estimation.} A source image $I_s$ is forwarded to a feature encoder to obtain a compact embedding $z_s$ and multi-scale features $f_s$. Then, $z_s$ and $f_s$ are processed by three decoders for different purposes, respectively 1) by a DepthDecoder to produce a set of depth maps $\tilde{D}_s$ (blue) at different resolutions,  2) by a second DepthDecoder, sharing weights with the previous one, after having been aligned to the target view according to $[R|t]_{s\rightarrow t}$, to produce a second set of depth maps $\tilde{D}_t$ (red), and 3) by the NVSDecoder, again after being aligned to the target viewpoint, to generate the novel view $\tilde{I}_t$. To align $z_s$ to the target view, we apply a latent space transformation (Sec. \ref{sec:latent}), while $f_s$ are warped by means of forward or backward warping, according to estimated depths $\tilde{D}_s$ and $\tilde{D}_t$ respectively (Sec. \ref{sec:warp}). }
        \label{fig:pipeline}
    \end{figure*}
    
    \section{Proposed Method}
    \label{Method}
    In this section, we introduce our framework designed to tackle the source-to-target NVS task by processing a single RGB image, while assisted by depth estimation as an auxiliary task. 
    We consider as input a 255$\times$255 RGB frame, i.e. the \textit{source} image $I_s$. During training, it is paired with a second, \textit{target} image -- i.e., the novel view $I_t$ we wish to generate starting from the source one -- and the relative transformation between the two, expressed by a 4$\times$4 matrix $[R|t]_{s\rightarrow t}$. The framework is trained to learn to generate the latter view, i.e., a view $\tilde{I}_t$ that closely resembles $I_t$. Synthesis is performed by aligning a compact latent space and the intermediate multi-scale features to the target viewpoint. Accordingly, we will refer to $\hat{\varepsilon}$ as the transformed counterpart of a generic $\varepsilon$ input from now on.
    We will now introduce our architecture, depicted in Fig. \ref{fig:pipeline}, for joint NVS and depth estimation, as well as the two key working principles implemented inside it.
    
    \subsection{Architecture}
    
    The source image $I_s$ is forwarded through an encoder-decoder pipeline to obtain an estimation of the view at target position $\tilde{I}_{t}$. The encoder reduces the resolution of the input image to a 1$\times$1 multidimensional tensor which embeds geometrical information, $z_s$, while a set of 5 multi-scale feature maps $f_s$ is obtained during this encoding process, respectively from half to $\frac{1}{32}$ of the original resolution. The generation step will then be performed by the NVSDecoder, the rightmost in Figure \ref{fig:pipeline}, which predicts the image from an embedding aligned with the desired viewpoints. 
    Such latent representation can be obtained, following \cite{chen2019mono, Hou_2021_WACV}, by directly manipulating the source embeddings, as detailed in the remainder.
    We used ResNet-18 convolutional blocks with pre-trained weights for the encoder, replacing the last pooling stage with an FC layer and a U-Net like architecture for the decoders.
    
    \subsection{Latent Space Transformation}
    \label{sec:latent}
    In order to decode an image from the compact embedding forwarded to the NVSDecoder, such an embedding should be geometrically aligned with the desired, target viewpoint. 
    Purposely, inspired by \cite{chen2019mono, Hou_2021_WACV} we train our encoder-decoder architecture to learn a compact latent representation that is equivariant to 3D transformations. This can be achieved by directly applying a 3D geometric transformation to the latent code itself. Specifically, the embedding $z_s$ is reshaped into a $N\times3$ structure and then multiplied by the transformation matrix $T_{s\rightarrow t} = [R|t]_{s\rightarrow t}$, that describes the roto-translation movement between source and target views.
    This produces a new latent code $\hat{z}_t$:
    \begin{equation}
    \hat{z}_t = z_s \times R_{s\rightarrow t} + t_{s \rightarrow t}
    \end{equation} 
    aligned to the viewpoint of the target image.
    This operation is fully differentiable, thus back-propagating through the whole framework will encourage the features encoder to extract embeddings meaningful to the 3D geometry of the scene. As a consequence, the transformed latent code can be used as a coarse 3D structure by the decoders to synthesize either the new view $\tilde{I}_t$ or its corresponding depth map $\tilde{D}_t$.
    
    \subsection{Direct and inverse warping}
    \label{sec:warp}
    While applying the 3D transformation in the latent space is sufficient for producing an embedding aligned with the target view, this is not enough for decoding accurate depth maps and novel views. Indeed, feeding the decoder with features coming from the encoder is crucial for preserving fine-grained details in the final results predicted by U-Net like architectures \cite{Ronneberger2015UNetCN}, and this is usually achieved by implementing skip connections. Therefore, to obtain features properly aligned with the target viewpoint, image-warping operators are used to establish a relationship between the pixel (homogeneous) coordinates $p_a$ of a generic image $I_a$ and those $p_b$ of a second frame $I_b$. This is achieved by using known intrinsics $K$, pose $[R|t]_{b\rightarrow a}$, and depth $\tilde{D}_a$.

    \begin{equation}
    p_b \sim K T_{a \rightarrow b} D_a K^{-1} p_a
    \end{equation}

    Accordingly, we can either apply inverse (backward) or direct (forward) warping depending on the features we want to warp, i.e., respectively from $p_b$ to $p_a$ or from $p_a$ to $p_b$. While the former is often implemented in the form of spatial transformer networks \cite{jaderberg2015spatial} for tasks such as self-supervised depth estimation \cite{Godard2017UnsupervisedMD}, 
    the latter is used less frequently, because of the collisions and holes it produces in the warped feature maps. As proposed in \cite{Hou_2021_WACV}, a single DepthDecoder -- i.e., the second in Figure \ref{fig:pipeline} -- would be sufficient to obtain depth map $\tilde{D}_t$, which can then be used to run inverse warping for $f_s$ and provide features $\hat{f}_t^{inv}$ to the NVSDecoder. However, the DepthDecoder itself cannot benefit from the encoder features, since they are still aligned with the source viewpoint. This produces sub-optimal depth predictions and, as a consequence, hinders the accuracy of the NVSDecoder.
    
    For this purpose, we deploy a second DepthDecoder -- i.e., the leftmost in Figure \ref{fig:pipeline} -- sharing the weight with the first one, to estimate a depth map $\tilde{D}_s$ from features $f_s$ and the embedding $z_s$. This first depth map, aligned with the source view, allows for running direct warping of the features $f_s$ and aligning them to the target viewpoint, obtaining $\hat{f}_t^{dir}$. The initial decoder benefiting of skip connections produces a more accurate depth map $\tilde{D}_t$, which in turn allows for a better warping of features $f_s$ into $\hat{f}_t^{inv}$ for the NVSDecoder and, finally, increases the quality of the generated, novel image.
    Since the features $f_s$ are extracted at different resolutions, intermediate depth maps estimated at the same resolutions are used both in direct and inverse warping operations.
    
    \subsection{Loss function}
    The whole model is trained in an end-to-end fashion, tightly linking NVS and depth estimation.
    The overall loss function is made of five terms detailed in the remainder.
    \textbf{Image reconstruction loss ($\mathcal{L}_{recon}$).}
    We supervise the NVSDecoder by upsampling each intermediate output $\tilde{I}_t^i$ from scale $i$ up to the original full resolution ($\tilde{I}_t^{i\uparrow}$) and by computing its L1 distance from ground truth image $I_t$.
    \begin{equation}
        \mathcal{L}_{recon} = \sum_{i=0}^\mathcal{S} |\tilde{I}_t^{i\uparrow} - I_t| \label{eq:recon}
    \end{equation}    
    \textbf{Photometric reprojection loss ($\mathcal{L}_{photo}$).} In addition to features warping, the second DepthDecoder also uses depth maps $\tilde{D}_t$ to warp source images $I_s$ and generate reprojected images $\hat{I_t}$, which provide self-supervision for the depth estimation task. Dissimilarity between the reprojected images $\hat{I_t}^i$ and the real images $I_t$ is measured at each intermediate depth map scale $i$, with the real images being downsampled accordingly.
    
    \begin{equation}
        \mathcal{L}_{photo} =\sum_{i=0}^\mathcal{S} |\hat{I}^{i}_t - I_t^{i\downarrow}| \label{eq:reproj}
    \end{equation}
    
    \textbf{VGG perceptual loss ($\mathcal{L}_{VGG}$)} As already demonstrated by \cite{Hou_2021_WACV}, the adoption of a VGG perceptual loss allows for enhancing the realism of generated results by increasing the sharpness of shapes. This is done by applying the feature extractor $\Theta$ of a pre-trained VGG16 network on the generated and the ground truth image individually, and then computing the L1 distance between such features.
    \begin{equation}
    \mathcal{L}_{VGG} = | \Theta(\tilde{I_t}) - \Theta(I_t) | \label{eq:vgg}
    \end{equation}
    
    \textbf{Edge-aware smoothness loss  ($\mathcal{L}_{smooth}$).} We further promote the smoothness of predicted depth maps by means of an edge-aware term, as proposed in \cite{Godard2017UnsupervisedMD, Godard2019DiggingIS}. 
    \begin{equation}
    \mathcal{L}^s_{smooth} = |\partial_{x} D_s|e^{\partial_{x} I_s} + |\partial_{y} D_s|e^{\partial_{y} I_s} \label{eq:smooth}
    \end{equation}
    This encourages smooth predictions for textureless regions of the images while preserving depth discontinuities. Similarly, $\mathcal{L}^t_{smooth}$ can be computed from $\tilde{D}_t$ and $I_t$.    
    
    \textbf{Depth consistency loss  ($\mathcal{L}_{skip}$).} In order to improve the results by the DepthDecoders, even when dealing with altered or incomplete features as a consequence of either latent space transformation or direct warping, an L1 loss is introduced. We compare the last depth map of the DepthDecoder processing $z_t$ and $f_t$, and the depth map from $\hat{z}_t$ and $\hat{f}_t^{dir}$ input. The first two terms are respectively embedding and features obtained by encoding the target view, and the last two terms by encoding the source view and applying the transformation introduced in Sec. \ref{sec:latent} and \ref{sec:warp} (i.e., step 1 in Figure \ref{fig:pipeline}).
    \begin{equation}
    \mathcal{L}_{skip} = |D(z_t,f_t) - D(\hat{z}_t,\hat{f}_t^{dir})| \label{eq:skip}
    \end{equation}
    The overall loss function is defined as 
    \begin{equation}
    \begin{split}
    \mathcal{L}_{tot} = &\alpha \mathcal{L}_{recon} + \beta \mathcal{L}_{photo} + \gamma \mathcal{L}_{VGG} + \\ &\delta (\mathcal{L}^s_{smooth}+\mathcal{L}^t_{smooth}) + \omega \mathcal{L}_{skip}  
    \end{split}
    \end{equation}
    with $\alpha$, $\beta$, $\gamma$, $\delta$, $\omega$ being weighting terms.
    
    \section{Experiments}
    We assess the effectiveness of our framework on both synthetic and real images and compare it with existing works built over the same principles, by using the pre-trained models released by the authors. We use the Adam optimizer with $\beta_1$ = 0.9, $\beta_2$ = 0.999, set the initial learning rate to $6e^{-5}$ with a linear decay program, and train our model for a total of 100k steps with a batch size of 8.
    The following sections introduce the datasets and metrics used for our evaluation, discuss the results in comparison with existing approaches, and present an ablation study to validate our design choices.
    
    \subsection{Dataset}
    
    Following \cite{Hou_2021_WACV,chen2019mono}, we evaluate our framework on ShapeNet\cite{Chang2015ShapeNetAI} and KITTI\cite{Geiger2012AreWR}, respectively representative of synthetic and real images.

    \textbf{ShapeNet.} It provides a collection of 3D synthetic objects with thousand of models for each category. We select two categories, in order to benchmark our framework on images with different topologies: chair models present complex designs, often with wiry and elongated features that are harder to be synthesized.
    On the opposite, car models are more consistent in the shape but the main body presents small details and rich textures that cannot be found in chairs. The dataset used is composed of 6777 chair models and 3514 car models, each rendered in 72 poses (both elevation and azimuth are taken with a sampling distance of 10°, respectively in [0°,40°] and [0°,360°] ranges) at a resolution of $512\times512$. For our model, those are downscaled to the proper input size and pairs of images are constructed in a range of [-40°,+40°] azimuth and random elevation. The train test split used is equivalent to the one used by \cite{Hou_2021_WACV}.

    \textbf{KITTI.} This dataset has been collected during driving sessions, using a car equipped with 
    standard cameras and Velodyne LiDAR sensors.
    It represents a standard real-world benchmark, used for many computer vision tasks. From KITTI we select 80k samples, obtained by excluding static scenes. The images are center-cropped to match the input size and pairs are constructed by taking the target view randomly within a distance of 7 frames from the source. We use the standard Eigen split\cite{Eigen2014DepthMP} for training, while we use the images from the Eigen test split as source views and sample corresponding target views from the sequences containing them. Poses are extracted from oxts files provided with the datasets.
    
    \subsection{Metrics}
    To quantitatively evaluate the results achieved by our framework, we adopt standard metrics for both tasks. Performances relative to NVS are expressed by the L1 norm, measuring the per-pixel difference between generated and ground truth images and the structural similarity index (SSIM) \cite{Wang2004ImageQA}, measuring the perceptual quality of the image. 
    On the KITTI dataset, we also measure the Peak Signal-to-Noise Ratio (PSNR), the higher the better, as well as the Learned Perceptual Image Patch Similarity (LPIPS) \cite{Zhang_2018_CVPR}.
    For the depth estimation task, we measure the accuracy of the predictions by the second DepthDecoder -- i.e., producing the depth for the target view, crucial for performing the target image synthesis -- with respect to ground truth depth, adopting a standard set of metrics from \cite{Eigen2014DepthMP} including square-root of scale-invariant logarithmic error (SILog), absolute relative error (Abs. Rel.), squared relative error (Sq. Rel.), root mean squared error (RMSE), root mean squared error between log depths (Log. RMSE) and threshold accuracy ($\delta < 1.25^i$, $i \in [1,2,3]$, with $\delta$ being the maximum between the prediction over ground truth ratio and its inverse).

\begin{table}[]
    \centering
    \caption{\textbf{Experimental results for NVS on ShapeNet.} We compare the synthesis quality achieved by Chen et al. \cite{chen2019mono}, Hou et al. \cite{Hou_2021_WACV} and our method. Best results in \textbf{bold}.}
    \label{tab:comparison-shapenet}
    \renewcommand{\tabcolsep}{15pt}
    \resizebox{0.5\textwidth}{!}{%
    \begin{tabular}{cl|ll}
    \hline 
       & & L1$\downarrow$ & SSIM$\uparrow$ \\ \hline
    \multirow{3}{*}{\rotatebox[origin=c]{90}{Chairs}} & Chen et al. \cite{chen2019mono} & 0.097          & 0.907 \\
    & Hou et al. \cite{Hou_2021_WACV} & 0.073          & 0.923 \\
    & Ours  & \textbf{0.026} & \textbf{0.943} \\ 
    \hline
    \multirow{3}{*}{\rotatebox[origin=c]{90}{Cars}} & Chen et al. \cite{chen2019mono} & 0.044 & 0.946 \\
    & Hou et al. \cite{Hou_2021_WACV} & 0.049 & 0.946 \\
    & Ours & \textbf{0.026} & \textbf{0.954} \\ \hline
    \end{tabular}%
    }
    \end{table}

\begin{figure}
    \centering
    \begin{tabular}{p{12mm} p{12mm} p{12mm} p{12mm} p{12mm}}
    \begin{subfigure}[b]{\subfigwidth}
        \caption{Source}
        \includegraphics[width=\subfigwidth,trim={1cm 1cm 1cm 1cm},clip]{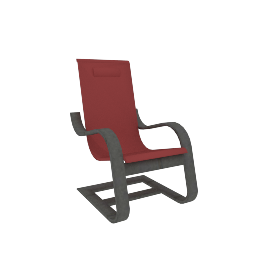}
    \end{subfigure} &
    \begin{subfigure}[b]{\subfigwidth}
        \caption{Target}
        \includegraphics[width=\subfigwidth,trim={1cm 1cm 1cm 1cm},clip]{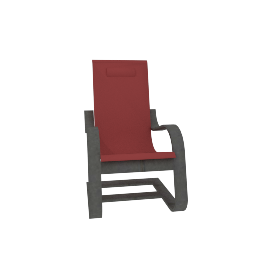}
    \end{subfigure} &
    \begin{subfigure}[b]{\subfigwidth}
        \caption{\cite{chen2019mono}}
        \includegraphics[width=\subfigwidth,trim={1cm 1cm 1cm 1cm},clip]{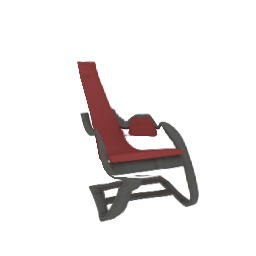}
    \end{subfigure} &
    \begin{subfigure}[b]{\subfigwidth}
        \caption{\cite{Hou_2021_WACV}}
        \includegraphics[width=\subfigwidth,trim={1cm 1cm 1cm 1cm},clip]{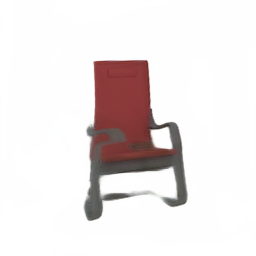}
    \end{subfigure} &
    \begin{subfigure}[b]{\subfigwidth}
        \caption{Ours}
        \includegraphics[width=\subfigwidth,trim={1cm 1cm 1cm 1cm},clip]{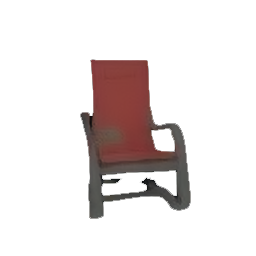}
    \end{subfigure} \\
    
    \includegraphics[width=\subfigwidth,trim={15mm 15mm 15mm 15mm},clip]{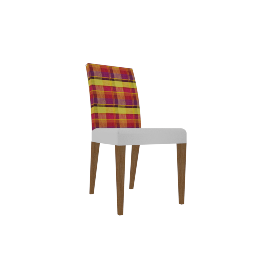} &
    \includegraphics[width=\subfigwidth,trim={15mm 15mm 15mm 15mm},clip]{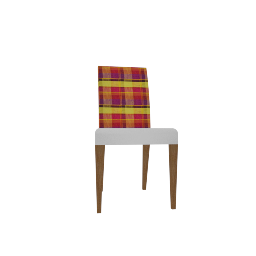} &
    \includegraphics[width=\subfigwidth,trim={15mm 15mm 15mm 15mm},clip]{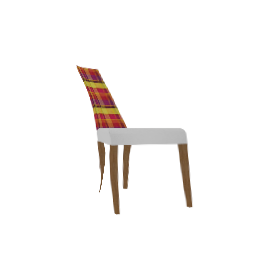} &
    \includegraphics[width=\subfigwidth,trim={15mm 15mm 15mm 15mm},clip]{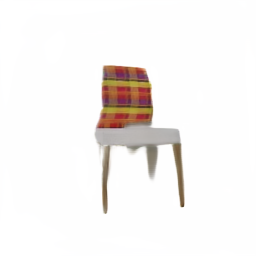} &
    \includegraphics[width=\subfigwidth,trim={15mm 15mm 15mm 15mm},clip]{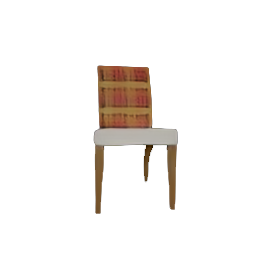} \\
    
    \includegraphics[width=\subfigwidth,trim={1cm 1cm 1cm 1cm},clip]{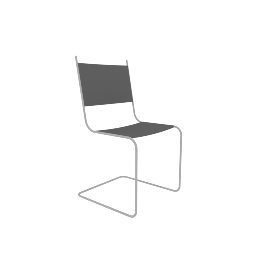} &
    \includegraphics[width=\subfigwidth,trim={1cm 1cm 1cm 1cm},clip]{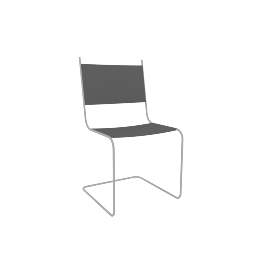} &
    \includegraphics[width=\subfigwidth,trim={1cm 1cm 1cm 1cm},clip]{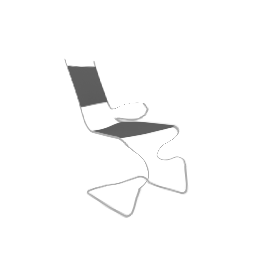} &
    \includegraphics[width=\subfigwidth,trim={1cm 1cm 1cm 1cm},clip]{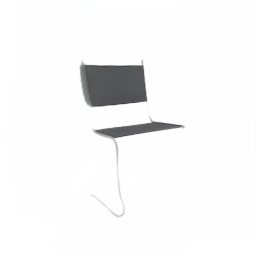} &
    \includegraphics[width=\subfigwidth,trim={1cm 1cm 1cm 1cm},clip]{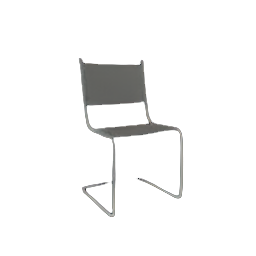} \\
    
    \includegraphics[width=\subfigwidth,trim={15mm 15mm 15mm 15mm},clip]{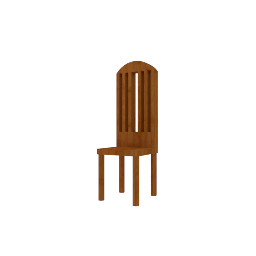} &
    \includegraphics[width=\subfigwidth,trim={15mm 15mm 15mm 15mm},clip]{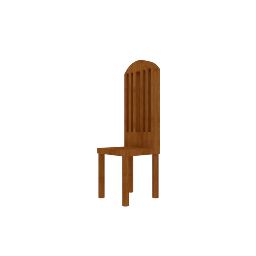} &
    \includegraphics[width=\subfigwidth,trim={15mm 15mm 15mm 15mm},clip]{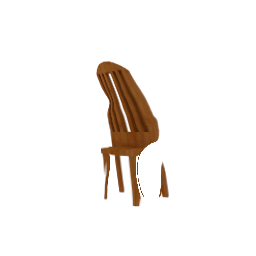} &
    \includegraphics[width=\subfigwidth,trim={15mm 15mm 15mm 15mm},clip]{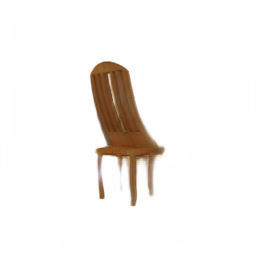} &
    \includegraphics[width=\subfigwidth,trim={15mm 15mm 15mm 15mm},clip]{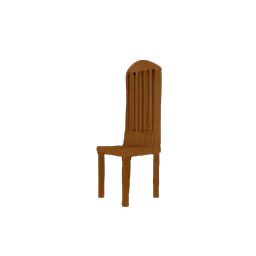} \\
    
    \includegraphics[width=\subfigwidth,trim={15mm 15mm 15mm 15mm},clip]{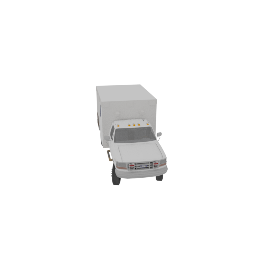} &
    \includegraphics[width=\subfigwidth,trim={15mm 15mm 15mm 15mm},clip]{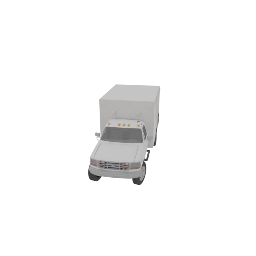} &
    \includegraphics[width=\subfigwidth,trim={15mm 15mm 15mm 15mm},clip]{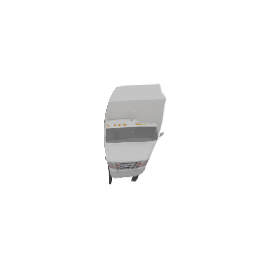} &
    \includegraphics[width=\subfigwidth,trim={15mm 15mm 15mm 15mm},clip]{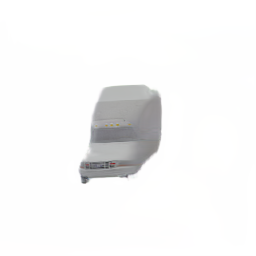} &
    \includegraphics[width=\subfigwidth,trim={15mm 15mm 15mm 15mm},clip]{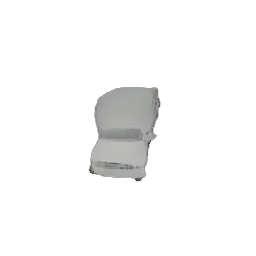} \\
    
    \includegraphics[width=\subfigwidth,trim={15mm 15mm 15mm 15mm},clip]{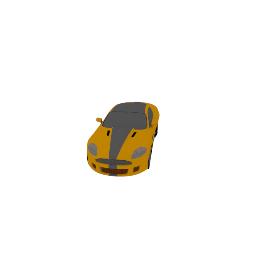} &
    \includegraphics[width=\subfigwidth,trim={15mm 15mm 15mm 15mm},clip]{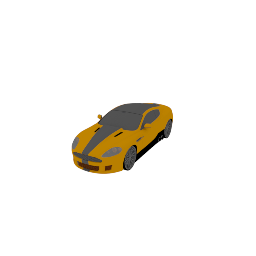} &
    \includegraphics[width=\subfigwidth,trim={15mm 15mm 15mm 15mm},clip]{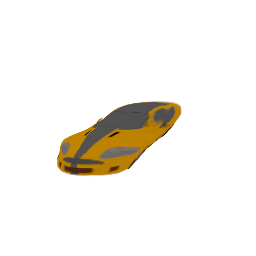} &
    \includegraphics[width=\subfigwidth,trim={15mm 15mm 15mm 15mm},clip]{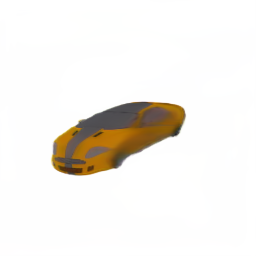} &
    \includegraphics[width=\subfigwidth,trim={15mm 15mm 15mm 15mm},clip]{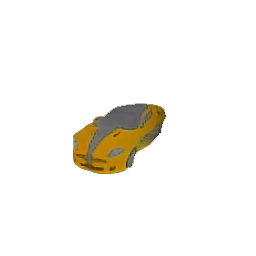} \\
    
    \includegraphics[width=\subfigwidth,trim={15mm 15mm 15mm 15mm},clip]{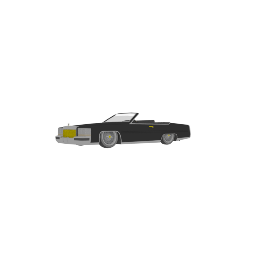} &
    \includegraphics[width=\subfigwidth,trim={15mm 15mm 15mm 15mm},clip]{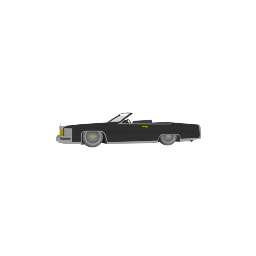} &
    \includegraphics[width=\subfigwidth,trim={15mm 15mm 15mm 15mm},clip]{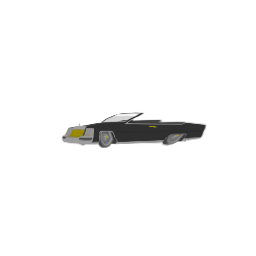} &
    \includegraphics[width=\subfigwidth,trim={15mm 15mm 15mm 15mm},clip]{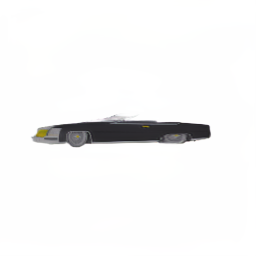} &
    \includegraphics[width=\subfigwidth,trim={15mm 15mm 15mm 15mm},clip]{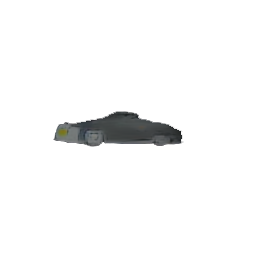} \\
    \end{tabular}
    \caption{\textbf{Qualitative results on ShapeNet.} From left to right: (a) source and (b) target images, novel views by (c) Chen et al. \cite{chen2019mono}, (d) Hou et al. \cite{Hou_2021_WACV} and (e) our framework. }
    \label{fig:comparison-shapenet}
    \end{figure}

    \begin{table*}[]
    \centering
    \caption{\textbf{Experimental results for NVS and depth estimation on KITTI.} We compare the synthesis quality (left) and depth accuracy (right) achieved by Chen et al., \cite{chen2019mono}, Hou et al. \cite{Hou_2021_WACV} and our method. Best results in \textbf{bold}.}
    \label{tab:comparison-kitti}
    \resizebox{\textwidth}{!}{%
    \begin{tabular}{l|llll|llllllll}
    \hline
    & L1$\downarrow$ & SSIM$\uparrow$ & PSNR$\uparrow$ & LPIPS$\downarrow$& SILog$\downarrow$ & Abs.Rel.$\downarrow$ & Sq.Rel.$\downarrow$ & RMSE$\downarrow$ & RMSE$_{log}\downarrow$ & $\delta_1\uparrow$ & $\delta_2\uparrow$ & $\delta_3\uparrow$ \\ 
    \hline
    Chen et al. \cite{chen2019mono} & 0.200          & 0.657          & 16.012          & 0.350          & 25.947            & 0.191                & 3.711               & 10.013           & 0.270                  & 0.755              & 0.889              & 0.944              \\
    Hou et al. \cite{Hou_2021_WACV} & 0.231          & 0.649          & 15.068          & 0.451          & 28.852            & 0.233                & 2.865               & 9.037            & 0.304                  & 0.633              & 0.848              & 0.939              \\
    Ours & \textbf{0.178} & \textbf{0.699}  & \textbf{17.248}  & \textbf{0.339} & \textbf{15.858}   & \textbf{0.116}       & \textbf{1.189}      & \textbf{6.089}   & \textbf{0.167}         & \textbf{0.863}     & \textbf{0.960}     & \textbf{0.988}     \\ \hline
    \end{tabular}%
    }
    \end{table*}

    \begin{figure*}
    \centering
    \begin{tabular}{p{15mm} p{15mm} p{15mm} p{15mm} p{15mm} p{15mm} p{15mm} p{15mm} p{15mm}}
    
    \begin{subfigure}[b]{\subfigwidthlarge}
        \caption{Source}
        \includegraphics[width=\subfigwidthlarge]{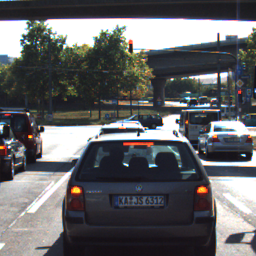}
    \end{subfigure} &
    \begin{subfigure}[b]{\subfigwidthlarge}
        \caption{Target}
        \includegraphics[width=\subfigwidthlarge]{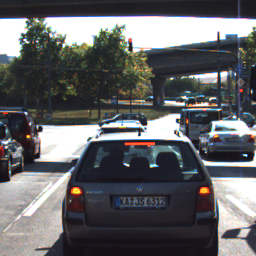}
    \end{subfigure} &
    \begin{subfigure}[b]{\subfigwidthlarge}
        \caption{\cite{chen2019mono}}
        \includegraphics[width=\subfigwidthlarge]{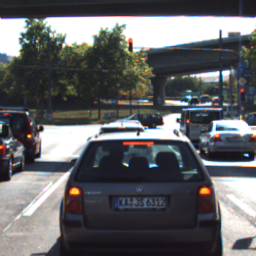}
    \end{subfigure} &
    \begin{subfigure}[b]{\subfigwidthlarge}
        \caption{\cite{Hou_2021_WACV}}
        \includegraphics[width=\subfigwidthlarge]{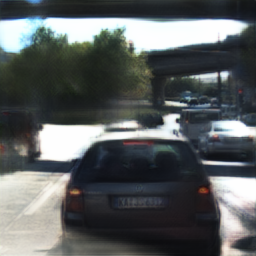}
    \end{subfigure} &
    \begin{subfigure}[b]{\subfigwidthlarge}
        \caption{Ours}
        \includegraphics[width=\subfigwidthlarge]{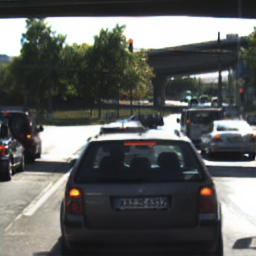}
    \end{subfigure} &
    \begin{subfigure}[b]{\subfigwidthlarge}
        \caption{GT}
        \includegraphics[width=\subfigwidthlarge]{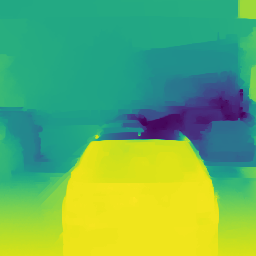}
    \end{subfigure} &
    \begin{subfigure}[b]{\subfigwidthlarge}
        \caption{\cite{chen2019mono}}
        \includegraphics[width=\subfigwidthlarge]{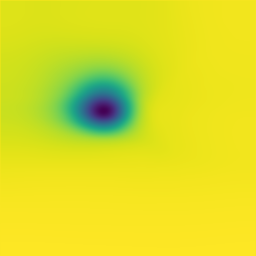}
    \end{subfigure} &
    \begin{subfigure}[b]{\subfigwidthlarge}
        \caption{\cite{Hou_2021_WACV}}
        \includegraphics[width=\subfigwidthlarge]{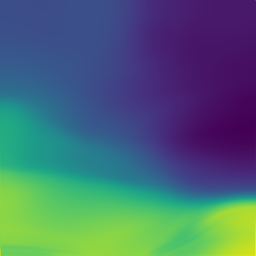}
    \end{subfigure} &
    \begin{subfigure}[b]{\subfigwidthlarge}
        \caption{Ours}
        \includegraphics[width=\subfigwidthlarge]{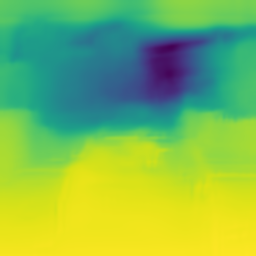}
    \end{subfigure} \\
    
    \includegraphics[width=\subfigwidthlarge]{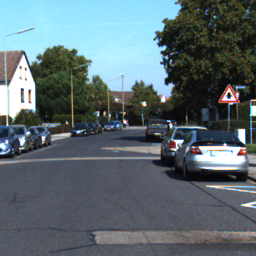} &
    \includegraphics[width=\subfigwidthlarge]{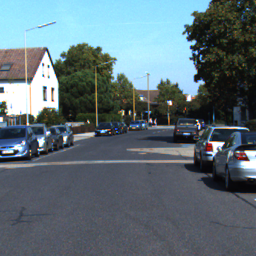} &
    \includegraphics[width=\subfigwidthlarge]{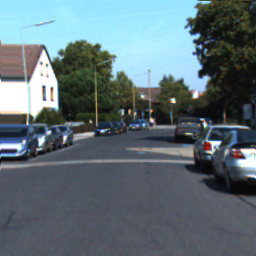} &
    \includegraphics[width=\subfigwidthlarge]{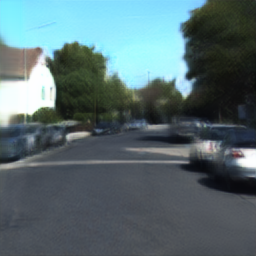} &
    \includegraphics[width=\subfigwidthlarge]{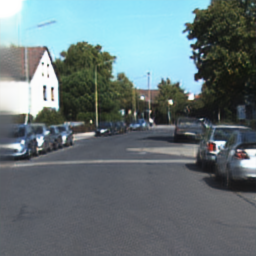} &
    \includegraphics[width=\subfigwidthlarge]{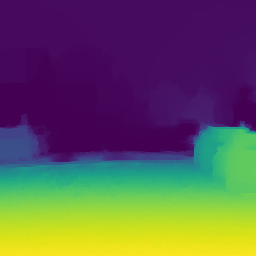} &
    \includegraphics[width=\subfigwidthlarge]{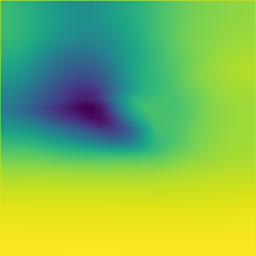} &
    \includegraphics[width=\subfigwidthlarge]{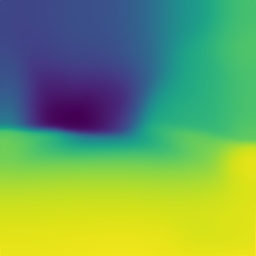} &
    \includegraphics[width=\subfigwidthlarge]{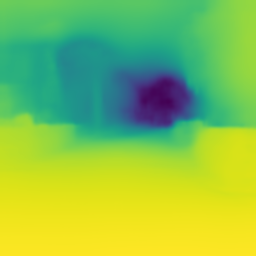} \\

    \includegraphics[width=\subfigwidthlarge]{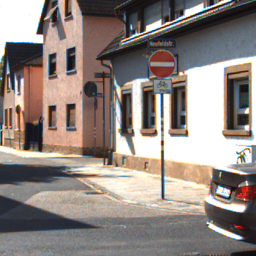} &
    \includegraphics[width=\subfigwidthlarge]{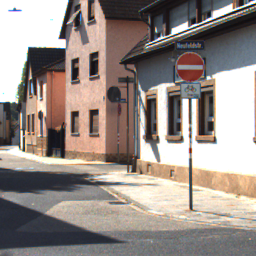} &
    \includegraphics[width=\subfigwidthlarge]{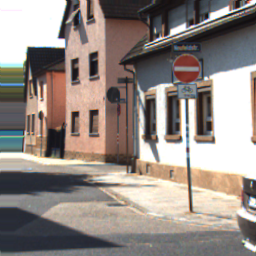} &
    \includegraphics[width=\subfigwidthlarge]{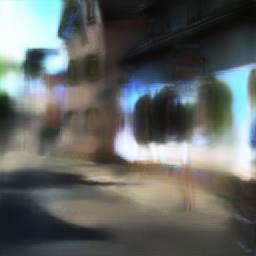} &
    \includegraphics[width=\subfigwidthlarge]{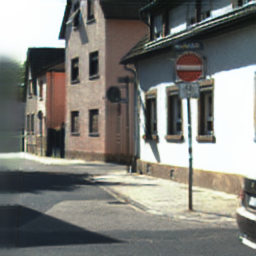} &
    \includegraphics[width=\subfigwidthlarge]{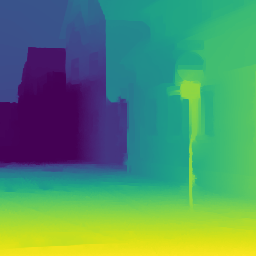} &
    \includegraphics[width=\subfigwidthlarge]{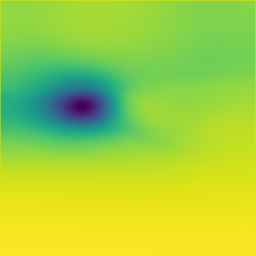} &
    \includegraphics[width=\subfigwidthlarge]{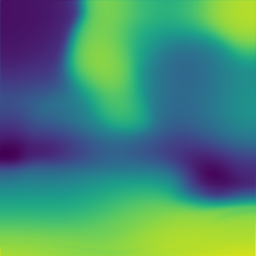} &
    \includegraphics[width=\subfigwidthlarge]{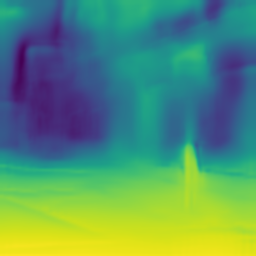} \\
    \end{tabular}
    \caption{\textbf{Qualitative results on KITTI.} From left to right: (a) source and (b) target images, novel views by (c) Chen et al. \cite{chen2019mono}, (d) Hou et al. \cite{Hou_2021_WACV} and (e) our framework, (f) ground truth depth, estimated depth maps by (g) Chen et al. \cite{chen2019mono}, (h) Hou et al. \cite{Hou_2021_WACV} and (i) our framework. }
    \label{fig:comparison-kitti}
    \end{figure*}

    \subsection{ShapeNet evaluation}
    Table \ref{tab:comparison-shapenet} shows results for NVS on the ShapeNet dataset, on chairs and cars.
    We can notice how our framework consistently achieves a higher quality synthesis with respect to existing works \cite{Hou_2021_WACV,chen2019mono}. This can be further appreciated qualitatively in Figure \ref{fig:comparison-shapenet}.
    On the one hand, we can notice how \cite{chen2019mono} often results in distorted shapes due to propagation of errors originated from a poor estimation of the target image depth, obtained exclusively by transformative projection in the synthesis step, while we can observe how \cite{Hou_2021_WACV} sometimes produces abnormally elongated shapes, mostly visible on cars.
    On the other hand, our improved encoding and decoding process allows for obtaining novel views geometrically consistent with the original one, preserving fine structures (e.g., chair legs).
    
    \subsection{KITTI evaluation}
    We further evaluate the effectiveness of our framework on KITTI, a real and more challenging dataset. Specifically, thanks to the availability of depth data collected by the Velodyne LiDAR sensor, we both evaluate the quality of the generated novel views, as well as the accuracy of estimated depth maps for such images. We point out how this depth estimation task differs from standard single-image depth estimation \cite{Godard2017UnsupervisedMD,Godard2019DiggingIS}, since depth is estimated according to a viewpoint different from the one of the source image, and thus is not directly comparable with this field of the literature.
    
    Table \ref{tab:comparison-kitti} collects quantitative comparisons between existing approaches  \cite{chen2019mono,Hou_2021_WACV} and our pipeline, reporting scores concerning both NVS and depth estimation tasks. We can notice how our proposal outperforms both on any metric.
    Figure \ref{fig:comparison-kitti} shows qualitative results concerning this experiment. We can notice how our framework, by jointly learning how to deal with both NVS and depth estimation, produces more detailed images and depth maps with respect to existing approaches.
    
    Finally, we also highlight a limitation common to any of the considered approaches, made evident by the KITTI dataset itself. By their formulation, these frameworks cannot properly deal with moving objects in the scene, often occurring in real environments such as those featured in the KITTI dataset. 
    As a consequence, rendering a target view in which we witness both a change of the camera viewpoint and the motion of some objects in the scene will produce a novel view in which the objects themselves did not move. Figure \ref{fig:comparison-kitti} shows an example on the third row. The rightmost car in the source image moves and, in addition to the small movement of the camera, disappears from the target view. However, each of the three methods generate images in which the car is still visible as if it was static. We leave explicitly modelling of the objects' movements in the scene as independent motions to future work.
    \begin{table*}[t]
    \centering
    \caption{\textbf{Ablation study on KITTI -- NVS and depth estimation.} We compare the synthesis quality (left) and depth accuracy (right) achieved by our full pipeline and ablated variants (I-VIII). Best results in \textbf{bold}, second best in \textcolor{blue}{blue}.}
    \label{tab:ablation}
    \resizebox{\textwidth}{!}{%
    \begin{tabular}{cl|cccc|cccccccc}
    \hline
                              & & {L1$\downarrow$} & {SSIM$\uparrow$} & {PSNR$\uparrow$} & {LPIPS$\downarrow$} & \multicolumn{1}{l}{SILog$\downarrow$} & \multicolumn{1}{l}{Abs.Rel.$\downarrow$}             & \multicolumn{1}{l}{Sq.Rel.$\downarrow$} & \multicolumn{1}{l}{RMSE$\downarrow$} & \multicolumn{1}{l}{RMSE$_{log}\downarrow$} & \multicolumn{1}{l}{$\delta_1\uparrow$} & \multicolumn{1}{l}{$\delta_2\uparrow$} & \multicolumn{1}{l}{$\delta_3\uparrow$} \\ \hline
    \rowcolor[HTML]{FFFFFF} 
    & Ours (full)           & {\color[HTML]{3166FF} 0.178}       & {\color[HTML]{333333} 0.699}       & {\color[HTML]{3166FF} 17.248}       & {\color[HTML]{333333} 0.339}       & {\color[HTML]{3166FF} 15.858}         & \cellcolor[HTML]{FFFFFF}{\color[HTML]{3166FF} 0.116} & {\color[HTML]{3166FF} 1.189}            & {\color[HTML]{3166FF} 6.089}         & {\color[HTML]{3166FF} 0.167}               & {\color[HTML]{3166FF} 0.863}           & \textbf{0.960}                        & \textbf{0.988}                          \\
    \hline
    (I) & No inverse warping        & 0.190                              & 0.688                              & 16.594                              & 0.404                              & 26.812                                & 0.217                                                & 2.665                                   & 8.498                                & 0.281                                      & 0.672                                  & 0.878                                 & 0.950                                   \\
    (II) & NVSDec w/o skips          & 0.309                              & 0.620                              & 12.843                              & 0.702                              & 21.497                                & 0.167                                                & 1.898                                   & 7.480                                & 0.224                                      & 0.776                                  & 0.927                                 & 0.974                                   \\
    (III) & DepthDec w/o skips        & 0.185                              & 0.690                              & 16.974                              & 0.352                              & 19.485                                & 0.151                                                & 1.669                                   & 7.112                                & 0.204                                      & 0.810                                  & 0.941                                 & 0.981                                   \\
    \hline
    (IV) & w/o $\mathcal{L}_{recon}$              & 0.189                              & 0.693                              & 17.015                              & {\color[HTML]{3166FF} 0.337}                              & \textbf{15.690}                       & \textbf{0.112}                                       & \textbf{1.163}                          & \textbf{6.057}                       & \textbf{0.165}                             & \textbf{0.868}                         & \textbf{0.960}                        & \textbf{0.988}                          \\
    (V) & w/o $\mathcal{L}_{VGG}$                & \textbf{0.172}                     & \textbf{0.709}                     & \textbf{17.364}                              & 0.427                              & 18.756                                & 0.136                                                & 1.480                                   & 6.833                                & 0.196                                      & 0.824                                  & 0.944                                 & 0.982                                   \\
    (VI) & w/o $\mathcal{L}_{photo}$              & 0.180                              & 0.698                              & 17.177                              & 0.338                              & 16.204                                & 0.117                                                & 1.361                                   & 6.326                                & 0.170                                      & 0.860                                  & 0.958                                 & 0.985                                   \\
    (VII) & w/o $\mathcal{L}_{smooth}$             & 0.183                              & 0.694                              & 17.064                              & 0.341                              & 16.189                                & 0.117                                                & 1.314                                   & 6.266                                & 0.170                                      & 0.861                                  & 0.958                                 & 0.986                                   \\
    (VIII) & w/o $\mathcal{L}_{consistency}$        & 0.179                              & {\color[HTML]{3166FF} 0.701}       & 17.160                              & \textbf{0.334}                              & 15.943                                & 0.117                                                & 1.260                                   & 6.148                                & 0.167                                      & {\color[HTML]{3166FF} 0.863}           & {\color[HTML]{3166FF} 0.959}          & {\color[HTML]{3166FF} 0.987}            \\ \hline
    \end{tabular}%
    }
    \end{table*}
    
    \subsection{Ablation study}
    \label{Ablation}
    We conclude our evaluation by measuring the impact of each component in our pipeline through an ablation study carried out on the KITTI dataset.
    Table \ref{tab:ablation} reports the outcome of this experiment, both in terms of NVS and depth estimation. Best results are reported in bold, while second bests are colored in blue. 
    
\begin{figure}[t]
    \centering
    \begin{tabular}{p{12mm} p{12mm} p{12mm} p{12mm} p{12mm} p{12mm} p{12mm} p{12mm} p{12mm}}
    
    \begin{subfigure}[b]{\subfigwidth}
        \caption{GT}
        \includegraphics[width=\subfigwidth]{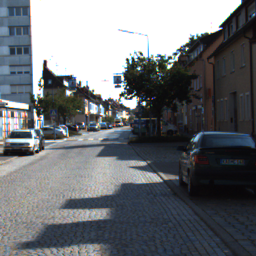}
    \end{subfigure} &
    \begin{subfigure}[b]{\subfigwidth}
        \caption{Full}
        \includegraphics[width=\subfigwidth]{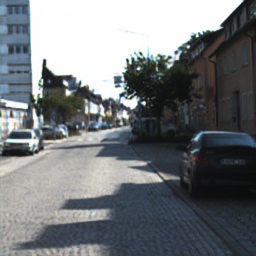}
    \end{subfigure} &
    \begin{subfigure}[b]{\subfigwidth}
        \caption{(I)}
        \includegraphics[width=\subfigwidth]{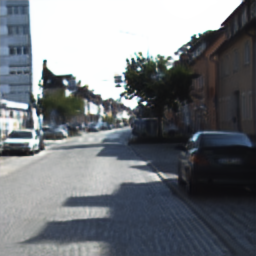}
    \end{subfigure} &
    \begin{subfigure}[b]{\subfigwidth}
        \caption{(II)}
        \includegraphics[width=\subfigwidth]{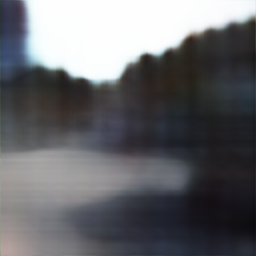}
    \end{subfigure} &
    \begin{subfigure}[b]{\subfigwidth}
        \caption{(III)}
        \includegraphics[width=\subfigwidth]{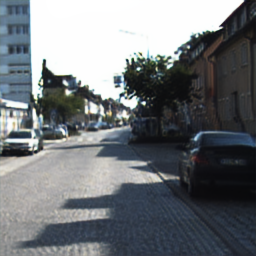}
    \end{subfigure} \\
    
    \includegraphics[width=\subfigwidth]{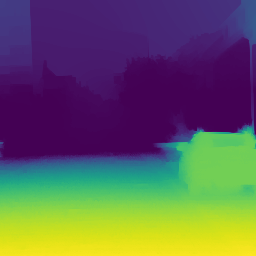} &    \includegraphics[width=\subfigwidth]{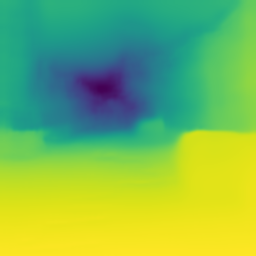} &
    \includegraphics[width=\subfigwidth]{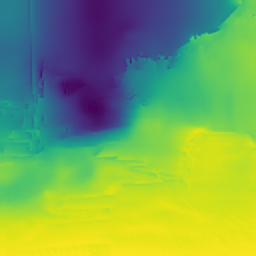} &
    \includegraphics[width=\subfigwidth]{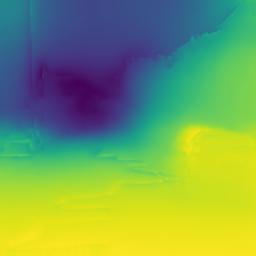} &
    \includegraphics[width=\subfigwidth]{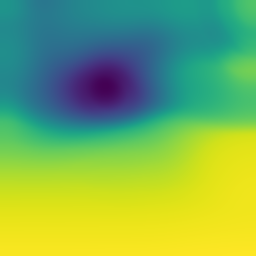} \\
    \end{tabular}
    \caption{\textbf{Qualitative results on KITTI -- ablation study.} From left to right: (a) ground truth target image (top) and ground truth depth (bottom), followed by (b-e) results achieved by configurations (I), (II) and (III) from Table \ref{tab:ablation}. }
    \label{fig:ablation}
    \end{figure}

    On top, we recall the results achieved by our full pipeline. Then, we first study the impact of architectural choices, reporting the results achieved by (I) forwarding direct warped features of the encoder to the NVSDecoder by means of source depth map -- i.e., by removing the second DepthDecoder -- as well as by (II) removing skip connections forwarded as input to the NVSDecoder or (III) the DepthDecoder.
    We can notice how dropping any of these design choices leads to a sensible decline in both NVS quality and predicted depth accuracy. Figure \ref{fig:ablation} confirms this qualitatively.
    Specifically, configuration (III) corresponds to \cite{Hou_2021_WACV}, indeed presenting the same blurring and distortion effects as it.
    Configuration (II) underline the clear effectiveness of the skip connections in the RGB space: here only a vague shape can be synthesized from the encoded information, in contrast with the highly detailed images obtained by our complete pipeline. Finally, removing skip connections directed to the DepthDecoders (III) affects the NVS process in a minor manner compared to what we observed for the NVSDecoder. However, the removal of such connections yields blurred depth maps, consequently causing the inverse warping to generate less effective reprojection of the encoder features to match the target viewpoint.
    
    At the bottom of the table, we also measure the impact of the different terms building up our loss function.
    We can notice how, by selectively turning off single terms, we can prioritize one task over the other.
    For instance, by neglecting the multiscale reconstruction loss $\mathcal{L}_{recon}$ (IV) we reduce the supervision given to NVS. Consequently, depth metrics improve at the expense of visual fidelity due to discoloration and artifacts. Removing the contribution of $\mathcal{L}_{VGG}$ (V), the network no longer explicitly optimizes for perceptual similarity in generated images, resulting in fuzzier outputs. However, surprisingly, this leads to an improvement in structural similarity, possibly due to the spreading of otherwise small artifacts in the sky of KITTI images.
    To conclude, we can notice how removing each one of the remaining terms independently (VI,VII,VIII) has minor consequences compared to the removal of $\mathcal{L}_{recon}$ and $\mathcal{L}_{VGG}$.

    \section{Conclusions}
    
    In this paper, we proposed a new pipeline for source-to-target novel view synthesis. Given a source image and a target viewpoint, our framework can generate a novel frame from this latter. Being our model explicitly trained to learn for both the image synthesis and depth estimation tasks, it achieves superior results both in terms of novel generated images as well as predicting their corresponding depth maps with respect to existing approaches for source-to-target view synthesis.
    Future research directions will aim at handling independently moving objects between the source and target images, making our approach suitable for novel view synthesis on unconstrained video sequences as well.

    \addtolength{\textheight}{0cm}
    
    \bibliographystyle{plain}
    \bibliography{refs}
    
    \end{document}